\documentclass{article}

\usepackage{times}
\usepackage{graphicx} %
\usepackage{subfigure} 

\usepackage{natbib}

\usepackage{algorithm}
\usepackage{algorithmic}
\usepackage{afterpage}

\usepackage{hyperref}

\usepackage[accepted]{icml2015}

\usepackage[table]{xcolor}
\usepackage{amsmath}
\usepackage{amssymb}
\usepackage{amsthm}
\usepackage{bbm}
\usepackage{enumerate}
\usepackage{xspace}
\usepackage{booktabs}

\usepackage{pascal-definitions}
\usepackage{hugo-definitions}

\newtheorem{thm}{Theorem}

\theoremstyle{definition}
\newtheorem{definition}[thm]{Definition}
\theoremstyle{plain}

\newcommand{\phib}{{\pmb \phi}}

\renewcommand{\ww}{{\mathbf{u}}}
\newcommand{\redplus}{``\red{$\boldsymbol{{+}}$}''}
\newcommand{\greenminus}{``\green{$\boldsymbol{{\pmb{-}}}$}''}

\icmltitlerunning{Domain-Adversarial Neural Networks}

\begin{document} 

\twocolumn[
\icmltitle{Domain-Adversarial Neural Networks}

\icmlauthor{Hana Ajakan$^1$}{hana.ajakan.1@ulaval.ca}
\icmlauthor{Pascal Germain$^1$}{pascal.germain@ift.ulaval.ca}
\icmlauthor{Hugo Larochelle$^2$}{hugo.larochelle@usherbrooke.ca}
\icmlauthor{Fran{\c c}ois Laviolette$^1$}{francois.laviolette@ift.ulaval.ca}
\icmlauthor{Mario Marchand$^1$}{mario.marchand@ift.ulaval.ca}
\icmladdress{${}^{1}$\,D{\'e}partement d'informatique et de g\'enie logiciel, Universit{\'e} Laval, Qu{\'e}bec, Canada\\
${}^{2}$\,D{\'e}partement d'informatique, Universit{\'e} de Sherbrooke, Qu{\'e}bec, Canada\\
All authors contributed equally to this work.}

\icmlkeywords{machine learning, domain adaptation, semi-supervised learning, neural networks}

\vskip 0.3in
]

\begin{abstract}
We introduce a new representation learning algorithm suited to the
context of domain adaptation, in which data at training and test time
come from similar but different distributions. Our algorithm is directly
inspired by theory on domain adaptation suggesting that, for effective domain
transfer to be achieved, predictions must be made based on a data
representation that cannot discriminate between the training (source)
and test (target) domains. We propose a training objective that
implements this idea in the context of a neural network, whose hidden
layer is trained to be predictive of the classification task, but
uninformative as to the domain of the input. Our experiments on
a sentiment analysis classification
benchmark, where the target domain data available at training time is unlabeled,
show that our neural network for domain adaption algorithm has
better performance than either a standard neural network or an SVM, even if trained on 
input features extracted with
the state-of-the-art marginalized stacked denoising autoencoders
of~\citet{Chen12}. %
\end{abstract} 

\section{Introduction}

The cost of generating labeled data for a new learning task is
often an obstacle for applying machine learning methods. 
There is thus
great incentive to develop ways of exploiting data from one problem
that generalizes to another.
Domain adaptation focuses on the situation where we have data
generated from two different, but somehow similar, distributions. One
example is in the context of sentiment analysis in written reviews,
where we might want to distinguish between the positive from the negative
ones. While we might have labeled data for reviews of one type of
products (\eg, movies), we might want to be able to generalize to
reviews of other products (\eg, books). Domain adaptation tries to achieve
such a transfer by exploiting an extra set of unlabeled training data
for the new problem to which we wish to generalize (\eg, unlabeled
reviews of books).

One of the main approach to achieve such a transfer is to learn
a classifier \emph{and} a representation which will 
favor the transfer. A large body of work exists on training
both a classifier and a representation that are linear \cite{BruzzoneM10S,pbda,CortesM14}.
However, recent research has shown that non-linear
neural networks can also be successful~\cite{Glorot+al-ICML-2011}.  Specifically, a
variant of the denoising autoencoder~\cite{VincentP2008}, known as marginalized stacked
denoising autoencoders (mSDA)~\cite{Chen12}, has demonstrated state-of-the-art
performance on this problem. By learning a representation
which is robust to input corruption noise, they have been able to 
learn a representation which is also more stable across changes of
domain and can thus allow cross-domain transfer.

In this paper, we propose to control the stability of representation
between domains explicitly into a neural network learning algorithm.  This approach is motivated by theory on domain
adaptation~\cite{BenDavid-NIPS06,BenDavid-MLJ2010} that suggests that
a good representation for cross-domain transfer is one for which an algorithm cannot learn to identify
the domain of origin of the input observation. We
show that this principle can be implemented into a neural network learning objective
that includes a term where the network's hidden layer
is working adversarially towards output connections
predicting domain membership.  The neural network is then simply trained by gradient descent  
on this objective. The success of this domain-adversarial neural network
(DANN) is confirmed by extensive experiments 
on both toy and real world datasets.
In particular, we show that DANN achieves better performances than a regular neural network and a SVM on a
sentiment analysis classification benchmark. 
Moreover, 
we show that DANN can reach state-of-the-art performance by
taking as input the representation learned by mSDA, confirming that minimizing domain discriminability
explicitly improves over on only relying on a representation which is
robust to~noise.

\section{Domain Adaptation}
\label{section:DA_theory}

We consider binary classification tasks where $\Xcal\subseteq\Rbb^n$ is the input space and $\Ycal\!\!=\!\{0,1\}$ is the label set.
Moreover, we have two different distributions over $\Xcal\!\times\! \Ycal$, called the {\it source domain} $\DS$ and the {\it target domain} $\DT$.
A \emph{domain adaptation} learning algorithm is then provided with a {\it labeled source sample} $S$ drawn {\it i.i.d.} from $\DS$, and an {\it unlabeled target sample} $T$ drawn {\it i.i.d.} from $\DTX$, where $\DTX$ is the marginal distribution of $\DT$ over $\Xcal$.
\begin{equation*}
S = \{(\xb^s_i,y^s_i)\}_{i=1}^{m} \sim (\DS)^m 
\,; \quad
T = \{\xb^t_i\}_{i=1}^{m'} \sim (\DTX)^{m'}\!.
\end{equation*}
The goal of the learning algorithm is to build a classifier $\eta:\Xcal\to\Ycal$ with a low \emph{target~risk}
\begin{equation*}
\RDT(\eta) \ \eqdef \Pr_{(\xb^t,y^t) \sim \DT} \left(\eta(\xb^t) \neq y^t\right),
\end{equation*}
while having no information about the labels of $\DT$.

\subsection{Domain Divergence}

To tackle the challenging domain adaptation task, many  approaches bound the target error by the sum of the source error and a notion of distance between the source and the target distributions. These methods are intuitively justified by a simple assumption: the source risk is expected to be a good indicator of the target risk when both distributions are similar. Several notions of distance have been proposed for domain adaptation~\cite{BenDavid-NIPS06,BenDavid-MLJ2010,Mansour-COLT09,MansourMR09,pbda}.
In this paper, we focus on the $\Hcal$-divergence used by ~\citet{BenDavid-NIPS06,BenDavid-MLJ2010}, and based on the earlier work of~\citet{kifer-2004}. %
\begin{definition}[\citet{BenDavid-NIPS06,BenDavid-MLJ2010, kifer-2004}]
Given two domain distributions $\DSX$ and $\DTX$ over~$\Xcal$, and a hypothesis class~$\Hcal$, the \emph{$\Hcal$-divergence} between $\DSX$ and $\DTX$ is
\begin{align*}
d_\Hcal(\DSX,\DTX)  \eqdef &  \\
&  \hspace{-18mm}
2 \sup_{\eta\in\Hcal} \,\bigg|\, 
\Pr_{\xb^s \sim \DSX} \big[\eta(\xb^s) = 1\big] - 
\Pr_{\xb^t \sim \DTX} \big[\eta(\xb^t) = 1\big]\,
\bigg|\,.
\end{align*}
\end{definition}
That is, the $\Hcal$-divergence relies on the capacity of the hypothesis class $\Hcal$ to distinguish between examples generated by $\DSX$ from examples generated by $\DTX$.
\citet{BenDavid-NIPS06,BenDavid-MLJ2010} proved that, for a symmetric hypothesis class $\Hcal$, one can compute the \emph{empirical $\Hcal$-divergence} between two samples $S\sim(\DSX)^m$ and  $T\sim(\DTX)^{m'}$ by computing
\begin{align} \label{eq:Hdiv_empirique}
 & \hat{d}_\Hcal(S,T) \ \eqdef  
  \\  
& 2\Bigg(\! 1\! -\! \min_{\eta\in\Hcal} \bigg[
\frac{1}{m} \sum_{i=1}^m I[\eta(\xb^s_i)\!=\!1] + \frac{1}{m'} \sum_{i=1}^{m'} I[\eta(\xb^t_i)\!=\!0]
\bigg]\! \Bigg),\nonumber
\end{align}
where $I[a]$ is the indicator function which is $1$ if predicate $a$ is true, and $0$ otherwise.

\subsection{Proxy Distance}
\label{section:PAD}

\citet{BenDavid-NIPS06,BenDavid-MLJ2010} suggested that, even if it is generally hard to compute $\hat{d}_\Hcal(S,T)$ exactly (\eg, when $\Hcal$ is the space of linear classifiers on $\Xcal$), we can easily approximate it by running a learning algorithm on the problem of discriminating between source and target examples. To do so, we construct a new dataset
\begin{equation}\label{eq:U}
U \ =\ \{(\xb^s_i, 1)\}_{i=1}^m \cup \{(\xb^t_i, 0)\}_{i=1}^{m'}\,,
\end{equation}
where the examples of the source sample are labeled $1$ and the examples of the target sample are labeled $0$. Then, the risk of the classifier trained on new dataset $U$ approximates the ``$\min$'' part of Equation~\eqref{eq:Hdiv_empirique}. 
Thus, given a test error~$\epsilon$ on the problem of discriminating between source and target examples, this \emph{Proxy A-distance} (PAD) is given by
\begin{equation} \label{eq:PAD}
\hat{d}_A \ = \ 2(1-2\epsilon)\,.
\end{equation}
In the experiments section of this paper, we compute the PAD value following the approach of \citet{Glorot+al-ICML-2011,Chen12}, \ie, we train a linear SVM on a subset of dataset $U$ (Equation~\eqref{eq:U}), and we use the obtained classifier error on the other subset as the value of~$\epsilon$ in Equation~\eqref{eq:PAD}.

\subsection{Generalization Bound on the Target Risk}
The work of \citet{BenDavid-NIPS06,BenDavid-MLJ2010} also showed that the $\Hcal$-divergence $d_\Hcal(\DSX,\DTX)$ is upper bounded by its empirical estimate $\hat{d}_\Hcal(S,T)$ plus a constant complexity term that depends on the \emph{VC dimension} of $\Hcal$ and the size of samples $S$ and $T$. By combining this result with a similar bound on the source risk, the following theorem is obtained.
\begin{thm}[\citet{BenDavid-NIPS06}] 
\label{thm:RDT_bound}
Let $\Hcal$ be a hypothesis class of VC dimension $d$.
With probability $1-\delta$ over the choice of samples $S\sim (\DS)^m$ and $T\sim (\DTX)^{m}$, for every $\eta\in\Hcal$:
\begin{align*}
\RDT(\eta) \ \leq& \ 
\RS(\eta) +  \sqrt{\frac{4}{m}\left( d \log\tfrac{2e\, m}{d}+  \log\tfrac{4}{\delta}\right) } 
\\ 
&{}+ \hat{d}_\Hcal(S,T) + 4\sqrt{ \frac{1}{m}\left(d \log\tfrac{2 m}{d}+  \log\tfrac{4}{\delta}\right) }
+ \beta\,,
\end{align*}
with $\beta \geq {\displaystyle\inf_{\eta^*\in\Hcal}} \left[ \RDS(\eta^*) + \RDT(\eta^*) \right]$\,, 
and 
\begin{equation*}
\RS(\eta) \ =\ \frac{1}{m}\dsum_{i=1}^m I\left[\eta(\xb^s_i) \neq y^s_i\right]
\end{equation*}
is the {empirical source~risk}.
\end{thm}
The previous result tells us that $\RDT(\eta)$ can be low only when the $\beta$ term is low, \ie, only when there exists a classifier that can achieve a low risk on both distributions. It also tells us that, to find a classifier with a small $\RDT(\eta)$ in a given class of fixed VC dimension, the learning algorithm should minimize (in that class) a trade-off between the source risk $\RS(\eta)$ and the empirical $\Hcal$-divergence $\hat{d}_\Hcal(S,T)$.  
As pointed-out by~\citet{BenDavid-NIPS06}, a strategy to control the $\Hcal$-divergence is to find a representation of the examples where both the source and the target domain are as indistinguishable as possible. Under such a representation, a hypothesis with a low source risk will, according to Theorem~\ref{thm:RDT_bound}, perform well on the target data.  
In this paper, we present an algorithm that directly exploits this idea.

\section{A Domain-Adversarial Neural Network} %
\label{section:dann}

The originality of our approach is to explicitly implement the idea exhibited by Theorem~\ref{thm:RDT_bound} 
into a neural network classifier.
That is,  to learn a
model that can generalize well from one domain to another, we ensure that
the internal representation of the neural network contains no discriminative information about the origin of the input (source or target), while preserving a low risk on the source (labeled) examples.

\subsection{Source Risk Minimization (Standard NN)}
\label{section:nn}

Let us consider the following standard neural network (NN) architecture with one hidden layer:
\begin{eqnarray}
\label{eq:hh}
\hh(\xx) &=& \sigm\big(\bb + \WW\xx\big) \,,\\
\ff(\xx) &=& \softmax\big(\cc + \VV\hh(\xx)\big)\,, \nonumber
\end{eqnarray}
with  
\vspace{-5mm}
\begin{eqnarray*}
\sigm(\aaa) &\eqdef& \left[\tfrac{1}{1+\exp(-a_i)}\right]_{i=1}^{|\aaa|},\\
\softmax(\aaa) &  \eqdef &  \left[\tfrac{\exp(a_i)}{\sum_{j=1}^{|\aaa|}\exp(a_j)}\right]_{i=1}^{|\aaa|}.
\end{eqnarray*}
Note that each component $f_y(\xx)$ of $\ff(\xx)$ denotes the conditional probability that the neural network assigns $\xx$ to class~$y$. Given a training source sample 
$S = \{(\xb^s_i,y^s_i)\}_{i=1}^{m}$,
the natural classification loss to use is the negative log-probability of the correct label:
\begin{equation*}
\Lcal\big(\ff(\xb), y\big) \ \eqdef \ \log\frac{1}{f_y(\xx)}\, .
\end{equation*}
This leads to the following learning problem on the source domain.
\begin{equation} \label{eq:loss}
\min_{\WW, \VV, \bb, \cc} \,\Bigg[ \frac{1}{m} \sum_{i=1}^m \Lcal \big(\ff(\xb_i^s), y_i^s\big)\Bigg]\,.
\end{equation}
We view the output of the hidden layer $\hh(\cdot)$ (Equation~\eqref{eq:hh}) as the internal representation of the neural network. 
Thus, we denote the source sample representations as 
$$\hh(S)\ \eqdef\  \big\{\hh(\xb^s_i)\big\}_{i=1}^m\,.$$

\subsection{A Domain Adaptation Regularizer}

Now, consider an unlabeled sample from the target domain
$T = \{\xb^t_i\}_{i=1}^{m'}\,$
and the corresponding representations \mbox{$\hh(T)\!=\!\{\hh(\xb^t_i)\}_{i=1}^{m'}$}. 
Based on Equation~\eqref{eq:Hdiv_empirique}, the empirical $\Hcal$-divergence of a symmetric hypothesis class $\Hcal$ between samples $\hh(S)$ and $\hh(T)$ is given by
\begin{align} \label{eq:Hdiv_hh}
& \hat{d}_\Hcal\big( \hh(S),\hh(T)\big) \,= \\
 \nonumber
& \mbox{\small$\displaystyle 2\bigg(\! 1 \!-\! \min_{\eta\in\Hcal}\! \bigg[
\frac{1}{m}\! \sum_{i=1}^m\! I\big[\eta(\hh(\xb^s_i))\! =\! 1\big] 
\!+\! \frac{1}{m'}\! \sum_{i=1}^{m'}\! I\big[\eta(\hh(\xb^t_i))\!=\!0\big]
\bigg] \!\Bigg)$}\!.
\end{align}
Let us consider $\Hcal$ as the class of hyperplanes in the representation space. Inspired by the Proxy A-distance (see Section~\ref{section:PAD}), we suggest estimating the ``$\min$'' part of Equation~\eqref{eq:Hdiv_hh} by a logistic regressor that model the probability that a given input (either $\xb^s$ or $\xb^t$)
is from the source domain $\DSX$ (denoted $z\!=\!1$) or the target domain $\DTX$ (denoted~$z\!=\!0$):
\begin{equation} \label{eq:o}
p(z=1\mid\phib) \ =\ o(\phib) \ \eqdef\ \sigm(d + \ww^\top \phib)\,,
\end{equation}
where $\phib$ is either $\hh(\xb^s)$ or $\hh(\xb^t)$. 
Hence, the function $o(\cdot)$ is a \emph{domain regressor}.

This enables us to add a domain adaptation term to the objective of Equation~\eqref{eq:loss}, giving 
the following problem to solve:
\begin{align}  \label{eqn:opt-domain}
&\min_{\WW, \VV, \bb, \cc} \!
\Bigg[ \frac{1}{m} \sum_{i=1}^m \Lcal \big(\ff(\xb_i^s), y_i^s\big)
\\
& \   +\! 
\lambda \max_{\ww, d} \!
\Bigg(
\!\!\!-\!\frac{1}{m}\! \sum_{i=1}^{m} \Lcal^d \big(o(\xb_i^s), 1\big) \!-\! \frac{1}{m'} \!\sum_{i=1}^{m'} \Lcal^d \big(o(\xb_i^t), 0\big)
\!\!\Bigg) \!\Bigg], \nonumber 
\end{align}
where the hyper-parameter $\lambda>0$ weights the domain adaptation regularization term and
\begin{align*}  
\Lcal^d \big(o(\xb), z\big)\ =\  
-z \log\big(o(\xb)\big) - (1\!-\!z) \log\big(1\!-\!o(\xb)\big)\,.
\end{align*}
In line with Theorem~\ref{thm:RDT_bound}, this optimization problem implements a trade-off between the minimization of the source risk $\RS(\cdot)$ and the divergence $\hat{d}_\Hcal( \cdot,\cdot)$. The hyper-parameter $\lambda$ is then used to tune the trade-off between these two quantities during the learning process.

\subsection{Learning Algorithm (DANN)}
\label{section:dann_algo}

\begin{figure} 
\centering
\includegraphics[width=.4\textwidth]{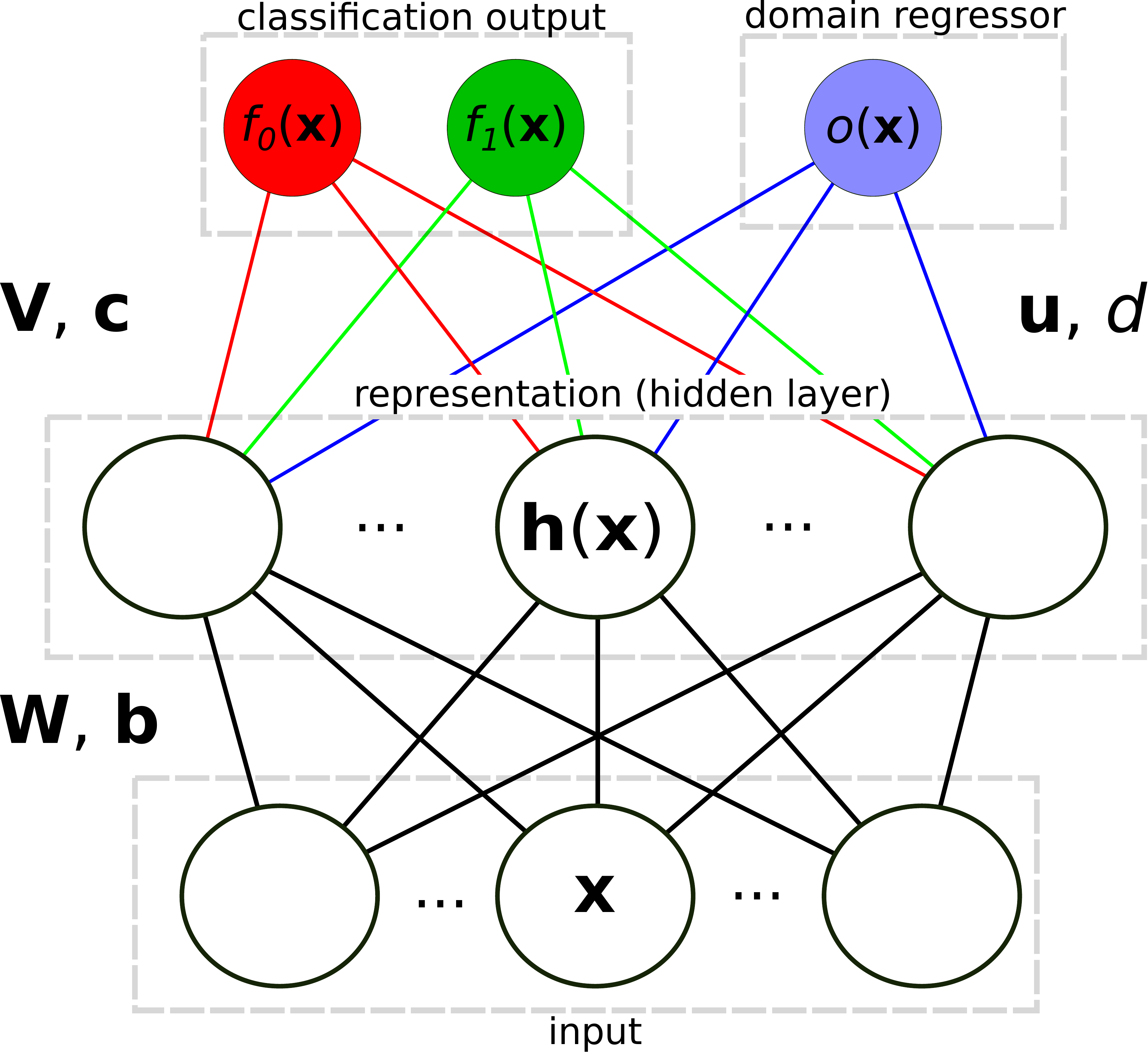}
\caption{DANN architecture \label{fig:DANN}}
\vspace{-2mm}
\end{figure}
We see that Equation~\eqref{eqn:opt-domain} involves a maximization operation. Hence, the
neural network (parametrized by $\{\WW, \VV, \bb, \cc\}$) and the
domain regressor (parametrized by $\{\ww, d\}$) are competing
against each other, in an adversarial way, for that term. 
The obtained \emph{domain adversarial neural network} (DANN) is illustrated by Figure~\ref{fig:DANN}. 
In DANN, the hidden layer $\hh(\cdot)$ maps an example (either source or target) into a representation in which the output layer $\ff(\cdot)$ accurately classifies the source sample, while the domain regressor $o(\cdot)$ is unable to detect if an example belongs to the source sample or the target sample.

To optimize Equation~\eqref{eqn:opt-domain}, one option would be to follow a hard-EM approach,
where we would alternate between optimizing until convergence the adversarial parameters $\ww, d$ and the
other regular neural network parameters $\WW, \VV, \bb, \cc$. However, we've found that a simpler stochastic gradient descent (SGD) approach is sufficient and works well in practice. Here, an SGD approach consists in sampling a pair of source and target example $\xb^s_i, \xb^t_j$ and updating a gradient step update of all parameters of DANN. Crucially, while the update of the regular parameters follows as usual the opposite direction of the gradient, for the adversarial parameters $\ww, d$ the step must follow the gradient's direction (since we maximize with respect to them, instead of minimizing). The algorithm is detailed in Algorithm~\ref{alg:stoch-up}. In the pseudocode, we use
$\ee(y)$ to represent a ``one-hot'' vector, consisting of all $0$s except for a $1$ at position~$y$. Also, $\odot$ is the element-wise product.
\begin{algorithm}[bt]
   \caption{DANN} %
   \label{alg:stoch-up}
\begin{algorithmic}[1]
{\small
   \STATE {\bfseries Input:} samples $S=\{(\xb^s_i, y^s_i)\}_{i=1}^m$ and $T=\{\xb^t_i\}_{i=1}^{m'}$,
   \STATE hidden layer size $l$, adaptation parameter $\lambda$, learning rate $\alpha$.
   \STATE {\bfseries Output:} neural network $\{\WW, \VV, \bb, \cc\}$ 
   \vspace{2mm}
   \STATE $\WW, \VV \leftarrow {\rm random\_init}(\,l\,)$
   \STATE $\bb, \cc, \ww, d \leftarrow 0$
   \WHILE{stopping criteria is not met}
   \FOR{$i$ from 1 to $m$}
   \STATE \# {\tt Forward propagation}
   \STATE $\hh(\xb^s_i) \leftarrow \sigm(\bb + \WW\xb^s_i)$
   \STATE $\ff(\xb^s_i) \leftarrow \softmax(\cc + \VV\hh(\xb^s_i))$
   \vspace{1.5mm}
   \STATE \# {\tt Backpropagation}
   \STATE $\Delta_{\cc} \leftarrow -(\ee(y^s_i)-\ff(\xb^s_i))$ 
   \STATE $\Delta_{\VV} \leftarrow \Delta_{\cc}~\hh(\xb^s_i)^\top$ 
   \STATE $\Delta_{\bb} \leftarrow \left(\VV^{\top} \Delta_{\cc}\right) \odot \hh(\xb^s_i) \odot (1-\hh(\xb^s_i))$
   \STATE $\Delta_{\WW} \leftarrow \Delta_{\bb} \cdot({\xb^s_i})^\top$
   \vspace{1.5mm}
   \STATE \# {\tt Domain adaptation regularizer...}
   \STATE \# {\tt ...from current domain}
   \STATE $o(\xb^s_i) \leftarrow \sigm(d + \ww^\top \hh(\xb^s_i))$
   \STATE $\Delta_d \leftarrow \lambda (1-o(\xb^s_i))$ ; $\Delta_{\ww} \leftarrow \lambda(1-o(\xb^s_i)) \hh(\xb^s_i)$
   \STATE ${\rm tmp} \leftarrow \lambda(1-o(\xb^s_i)) \ww \odot \hh(\xb^s_i) \odot (1-\hh(\xb^s_i))$
   \STATE $\Delta_{\bb} \leftarrow \Delta_{\bb} + {\rm tmp}$ ; $\Delta_{\WW} \leftarrow \Delta_{\WW} + {\rm tmp}\cdot({\xb^s_i})^\top$
   \label{algoline:omit1}
      \vspace{1.5mm}
   \STATE \# {\tt ...from other domain}
   \STATE $j \leftarrow {\rm uniform\_integer}(1,\dots,m')$
   \STATE $\hh(\xb^t_j) \leftarrow \sigm(\bb + \WW\xb^t_j)$
   \STATE $o(\xb^t_j) \leftarrow \sigm(d + \ww^\top \hh(\xb^t_j))$ 
   \STATE $\Delta_d \leftarrow \Delta_d - \lambda o(\xb^t_j)$; $\Delta_{\ww} \leftarrow \Delta_{\ww} - \lambda o(\xb^t_j) \hh(\xb^t_j)$
   \STATE ${\rm tmp} \leftarrow -\lambda o(\xb^t_j) \ww \odot \hh(\xb^t_j) \odot (1-\hh(\xb^t_j))$
   \STATE $\Delta_{\bb} \leftarrow \Delta_{\bb} + {\rm tmp}$ ; $\Delta_{\WW} \leftarrow \Delta_{\WW} + {\rm tmp}\cdot(\xb^t_j)^\top$
   \label{algoline:omit2}
      \vspace{1.5mm}
   \STATE \# {\tt Update neural network parameters}
   \STATE $\WW \leftarrow \WW - \alpha \Delta_{\WW}$ ; $\VV \leftarrow \VV - \alpha \Delta_{\VV}$
   \STATE $\bb \leftarrow \bb - \alpha \Delta_{\bb}$ ; $\cc \leftarrow \cc - \alpha \Delta_{\cc}$
      \vspace{1.5mm}
   \STATE \# {\tt Update domain classifier parameters}
   \STATE $\ww \leftarrow \ww + \alpha \Delta_{\ww}$ ; $d \leftarrow d + \alpha \Delta_{d}$
   \label{algoline:da}
      \ENDFOR
   \ENDWHILE
   }  \vspace{-1mm}
\end{algorithmic}
\end{algorithm}
For each experiment described in this paper, we used \emph{early stopping} as the stopping criteria: we split the source labeled sample to use $90\%$ as the training set $S$ and the remaining $10\%$ as a validation set $S_V$. We stop the learning process when the risk on $S_V$ is minimal.

\section{Related Work}

As mentioned previously, the general approach of achieving domain 
adaptation by learning a new data representation has been
explored under many facets. A large part of the
literature however has focused mainly on linear 
hypothesis
\citep[see for instance][]{BlitzerMP06,BruzzoneM10S,pbda,BaktashmotlaghM2013,CortesM14}.
More recently, non-linear representations have become increasingly studied,
including neural network representations~\cite{Glorot+al-ICML-2011,LiY2014} and most notably
the state-of-the-art mSDA~\cite{Chen12}. That literature has mostly focused
on exploiting the principle of robust representations, based on the denoising autoencoder paradigm~\cite{VincentP2008}. 
One of the contribution of this work is to show that domain discriminability is another principle
that is complimentary to robustness and can improve cross-domain adaptation.

What distinguishes this work from most of the domain adaptation literature is DANN's inspiration
from the theoretical work of \citet{BenDavid-NIPS06,BenDavid-MLJ2010}. Indeed, DANN directly optimizes
the notion of $\Hcal$-divergence. We do note the work of \citet{HuangY12},
in which HMM representations are learned for word tagging using a posterior regularizer that
is also inspired by \citeauthor{BenDavid-MLJ2010}'s work. In addition to the tasks being different (word tagging versus
sentiment classification), we would argue that DANN learning objective more closely optimizes
the $\Hcal$-divergence, with \citet{HuangY12} relying on cruder approximations 
for efficiency reasons.

The idea of learning representations that are indiscriminate of some auxiliary label has
also been explored in other contexts. For instance, \citet{ZemelR2013} proposes the notion of fair representations,
which are indiscriminate to whether an example belongs to some identified set of groups. The resulting
algorithm is different from DANN and not directly derived from the $\Hcal$-divergence. 

Finally, we mention some related research on using an adversarial (minimax) formulation to learn a model, such as a classifier or a neural network, from data.
There has been work on learning linear classifiers that are robust to changes in the input distribution, based on a minimax formulation~\cite{BagnellD2005,LiuA2014}. This work however assumes that a good feature representation of the input for a linear classifier is available and doesn't address the problem of learning it.
We also note the work of \citet{GoodfellowI2014}, who propose generative adversarial networks to learn a good generative model of the true data distribution. This work shares with DANN the use of an adversarial objective, applying it instead to the unsupervised problem of generative modeling.

\section{Experiments}

\subsection{Toy Problem}
\label{sec:toy_problem}

\begin{figure*}[t]
\centering
\subfigure[Standard NN and a \emph{non adverserial} domain  regressor (Algorithm~\ref{alg:stoch-up}, without Lines~\ref{algoline:omit1} and~\ref{algoline:omit2})]
{\footnotesize\bf\sc
\begin{minipage}{.245\textwidth}\centering
Label classification\\[1mm]
\includegraphics[width=1\textwidth]{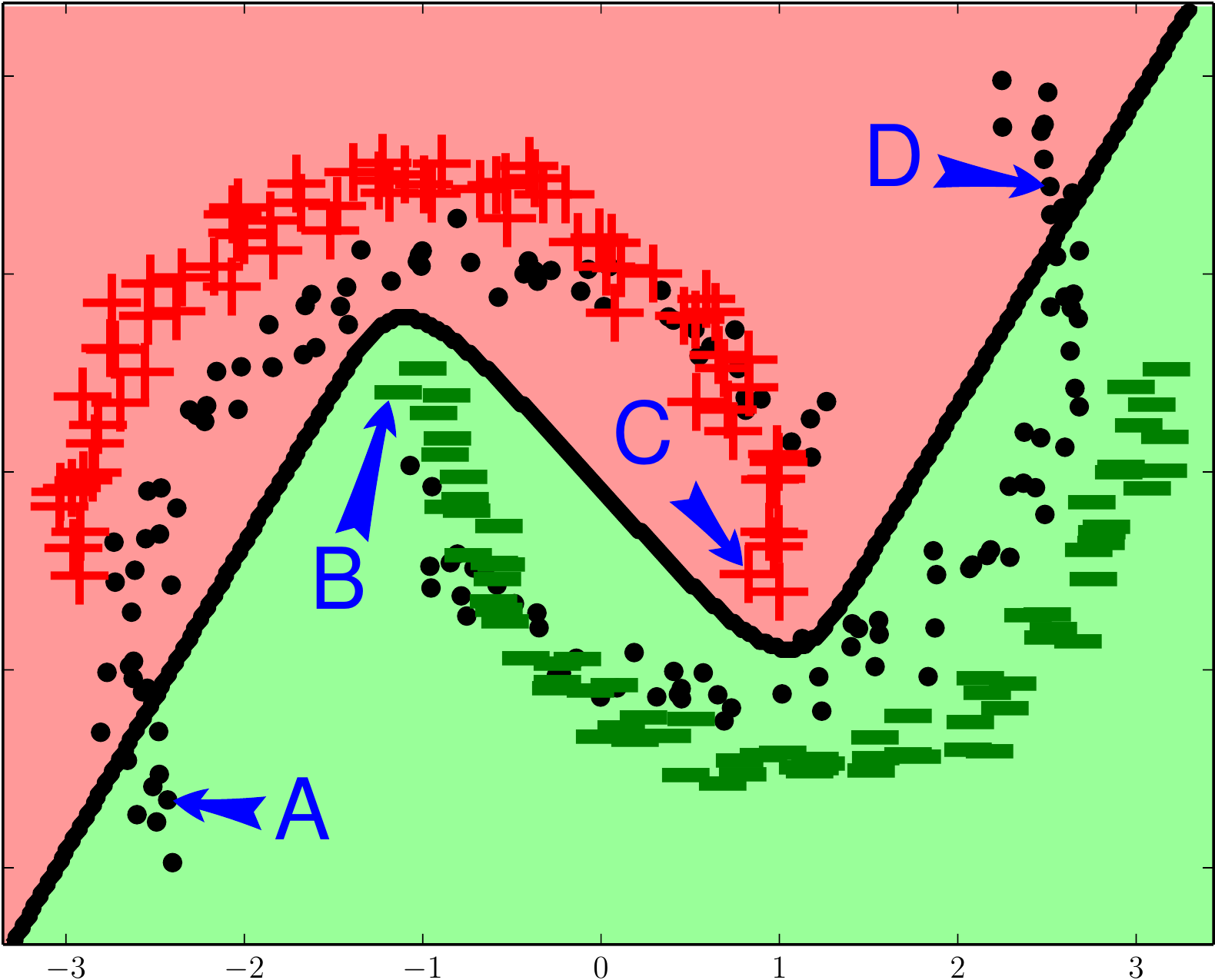}
\vspace{-1mm}
\end{minipage}
\begin{minipage}{.245\textwidth}\centering
Representation PCA\\[1mm]
\includegraphics[width=1\textwidth]{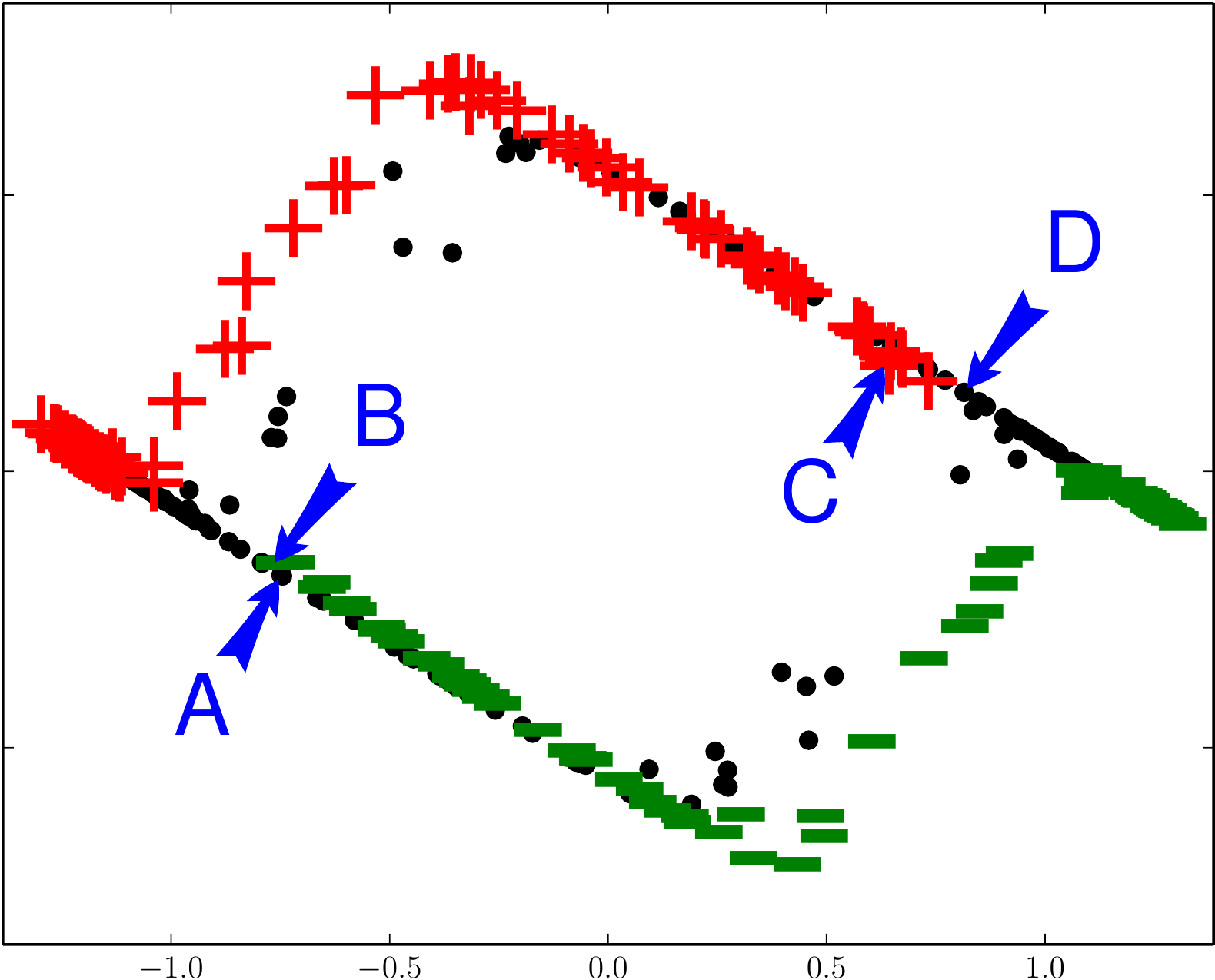}
\vspace{-1mm}
\end{minipage}
\begin{minipage}{.245\textwidth}\centering
Domain classification\\[1mm]
\includegraphics[width=1\textwidth]{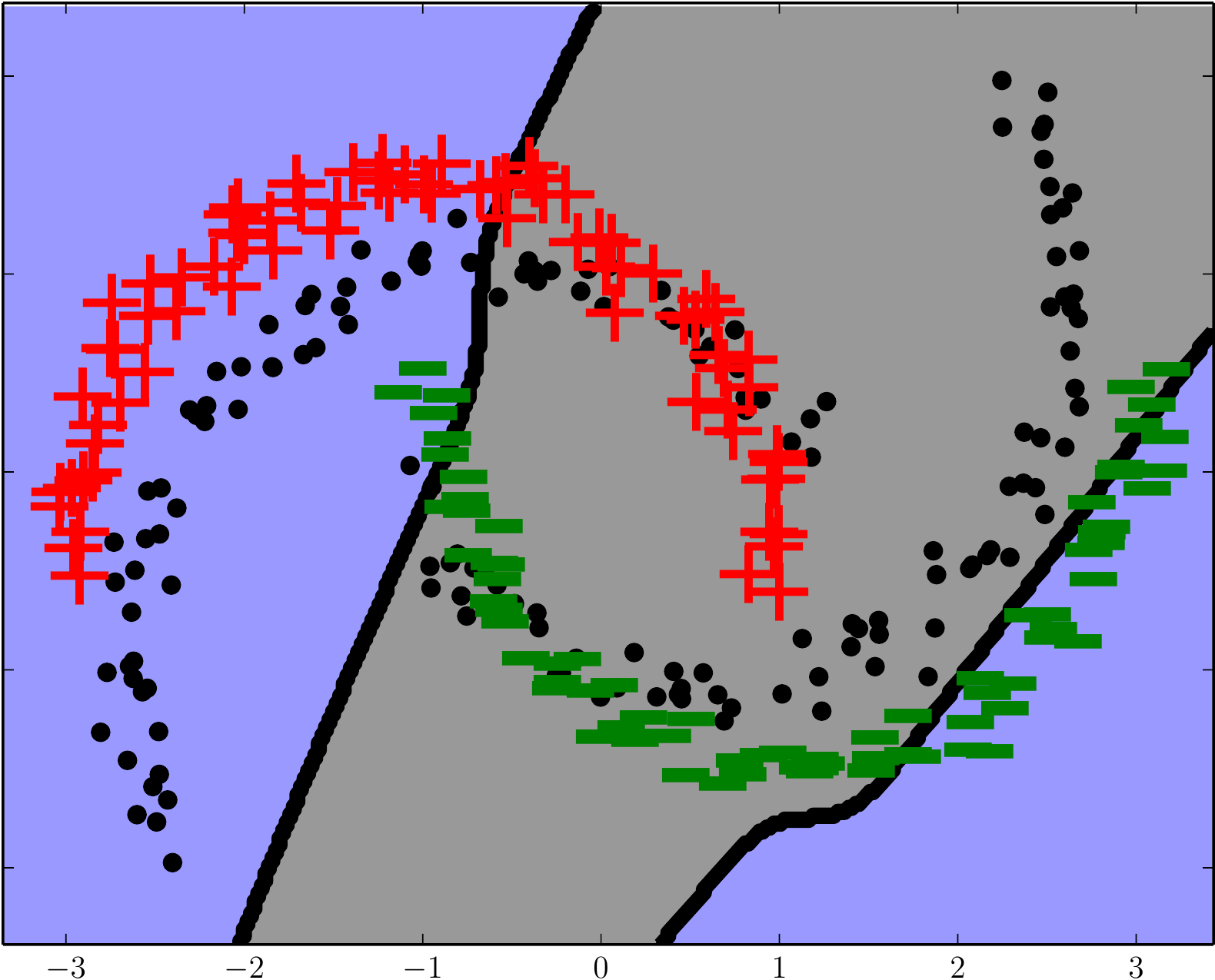}
\vspace{-1mm}
\end{minipage}
\begin{minipage}{.245\textwidth}\centering
Hidden neurons\\[1mm]
\includegraphics[width=1\textwidth]{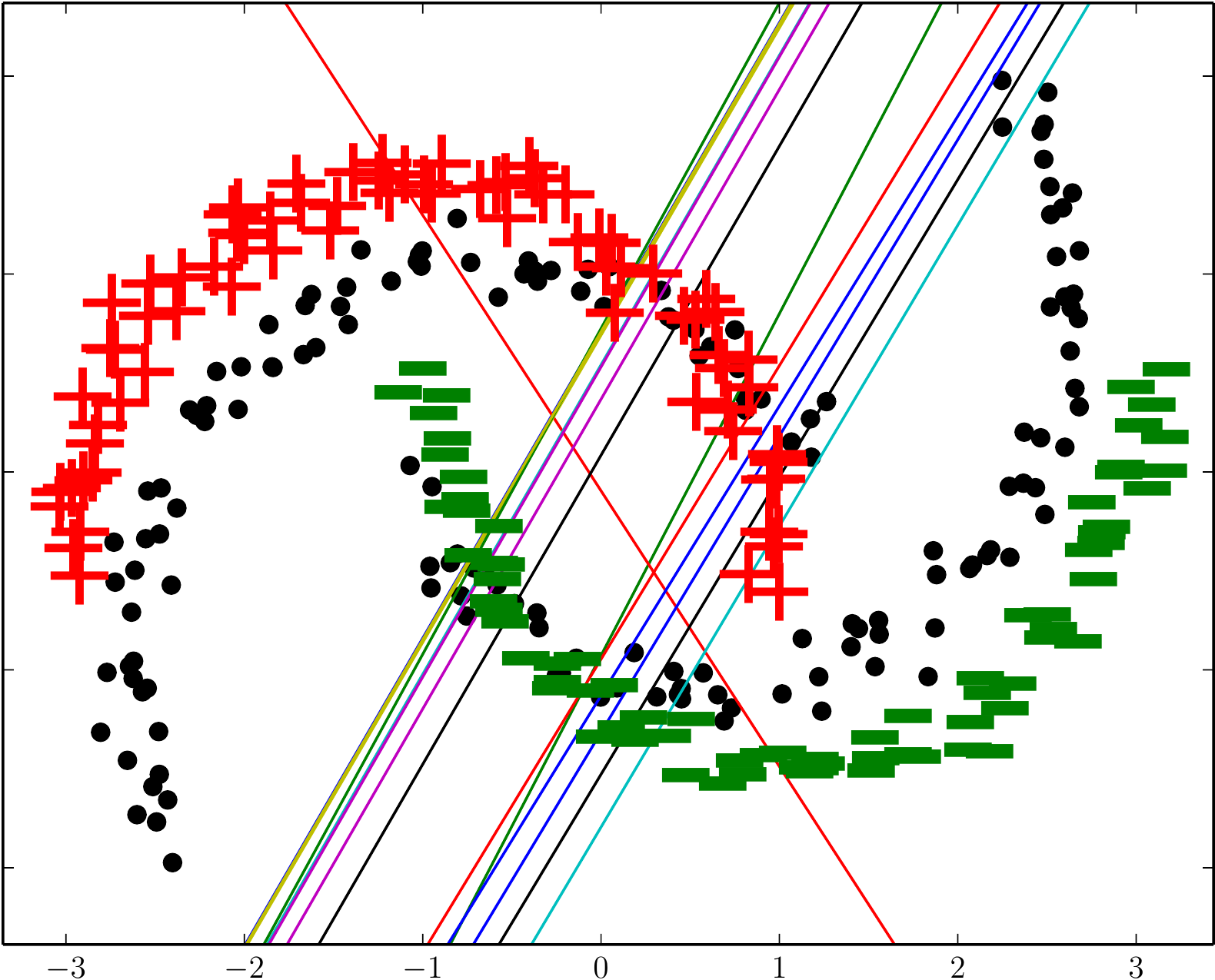}
\vspace{-1mm}
\end{minipage} 
\label{fig:2moons_NN}
 }
 \subfigure[DANN (Algorithm~\ref{alg:stoch-up})]
 {\footnotesize\bf\sc
 \begin{minipage}{.245\textwidth}\centering
 \includegraphics[width=1\textwidth]{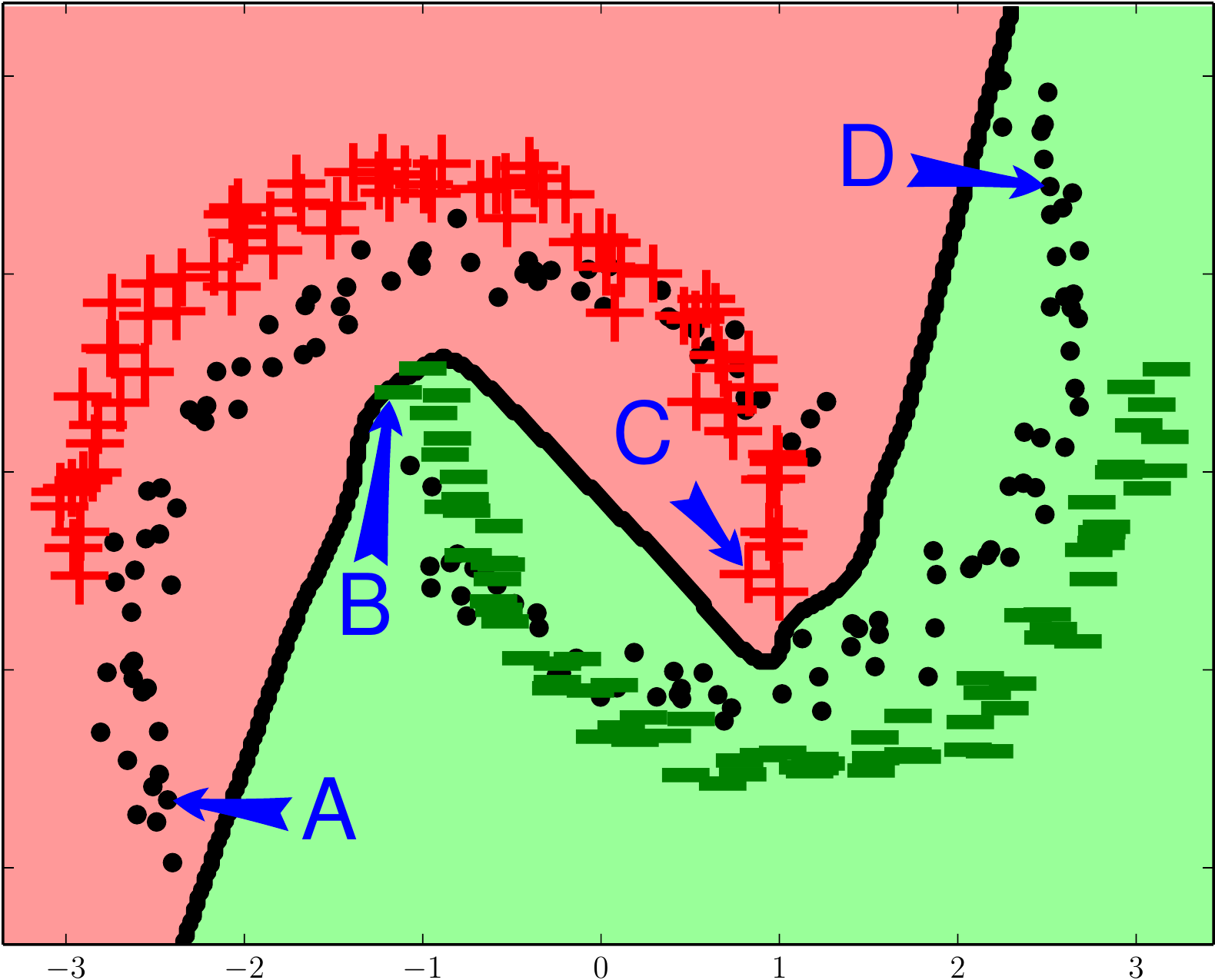}
 \vspace{-1mm}
 \end{minipage}
 \begin{minipage}{.245\textwidth}\centering
 \includegraphics[width=1\textwidth]{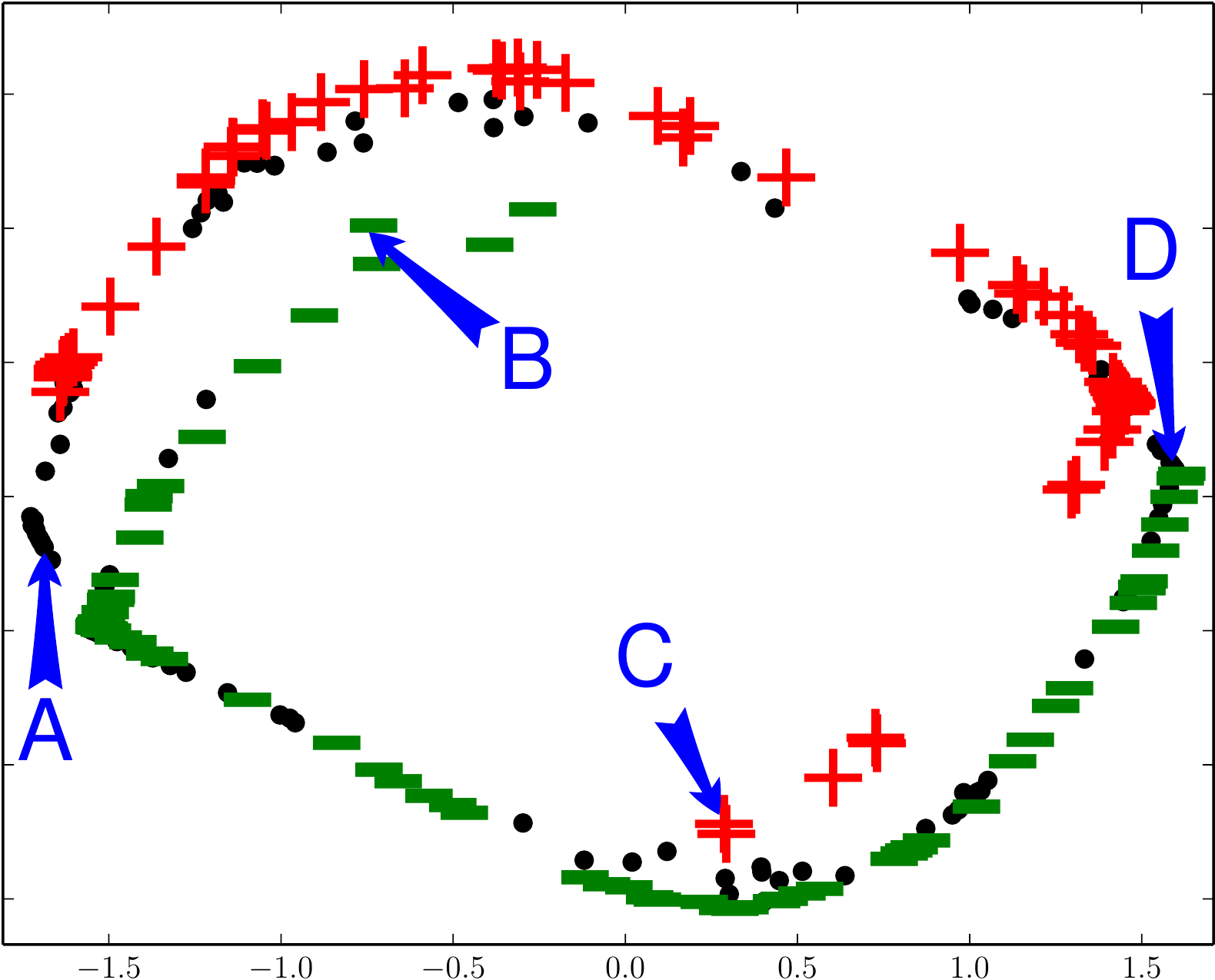}
 \vspace{-1mm}
 \end{minipage}
 \begin{minipage}{.245\textwidth}\centering
 \includegraphics[width=1\textwidth]{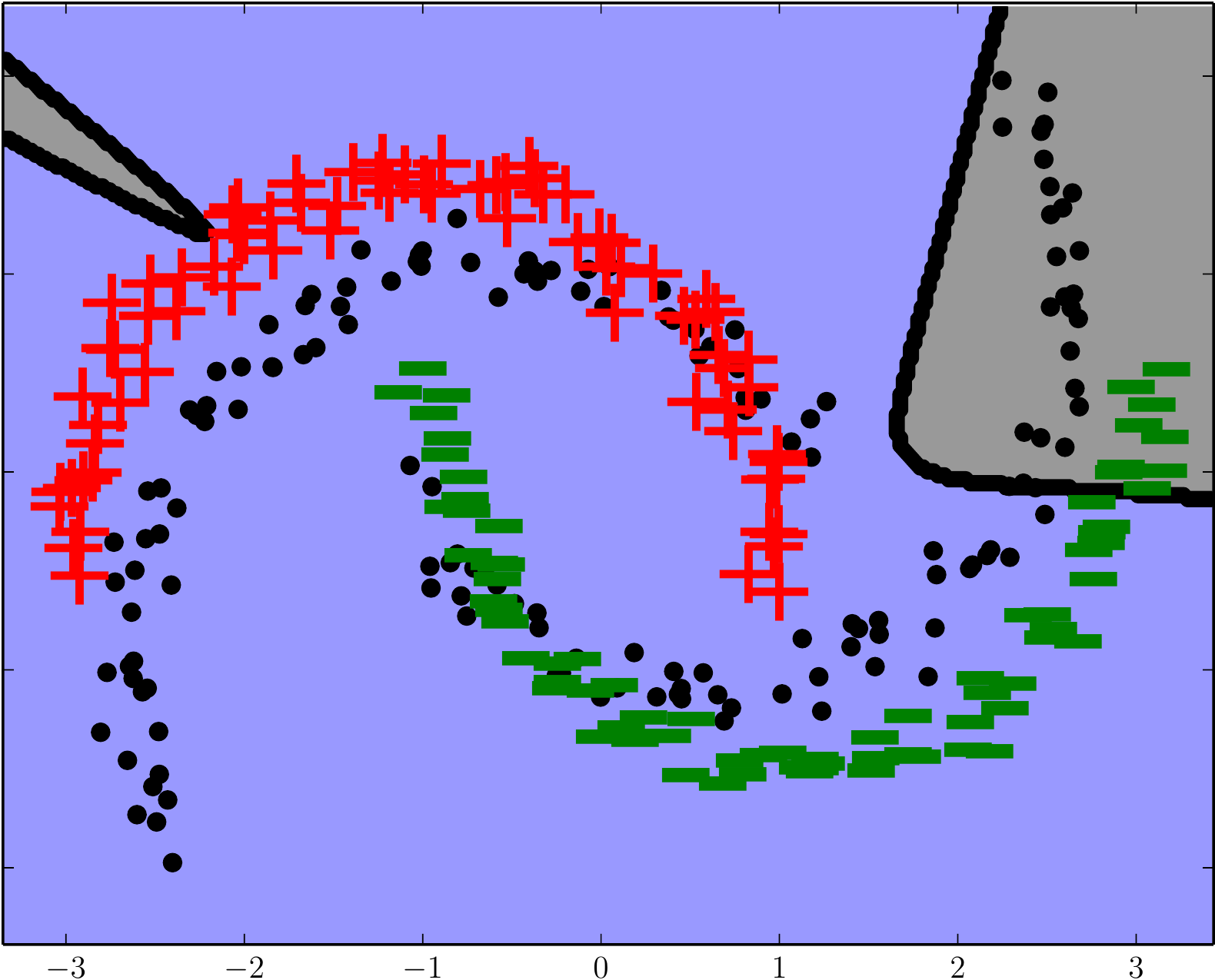}
 \vspace{-1mm}
 \end{minipage}
 \begin{minipage}{.245\textwidth}\centering
 \includegraphics[width=1\textwidth]{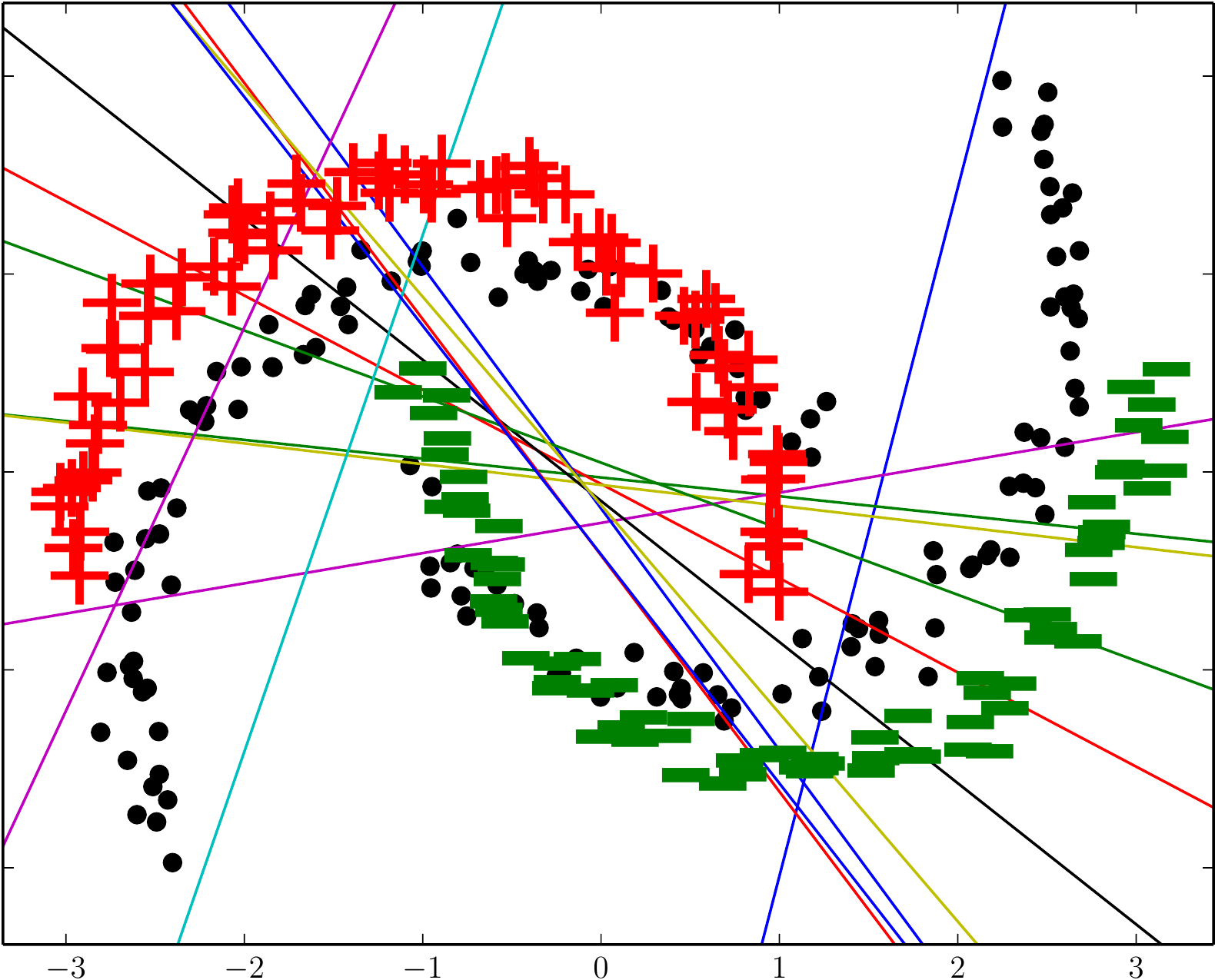}
 \vspace{-1mm}
 \end{minipage}
 \label{fig:2moons_DANN}
  }
\caption{The \emph{inter-twinning moons} toy problem. 
We adopt the colors of Figure~\ref{fig:DANN} for classification output and domain regressor value.
\label{fig:2moons} }
\end{figure*}

As a first experiment, we study the behavior of the proposed DANN algorithm on a variant of the \emph{inter-twinning moons} 2D problem, where the target distribution is a rotation of the source distribution. For the source sample $S$, we generate a lower moon and an upper moon labeled $0$ and $1$ respectively, each of which containing $150$ examples.  The target sample $T$ is obtained by generating a sample in the same way as $S$ (without keeping the labels) and then by rotating each example by $35\degree$. Thus, $T$ contains $300$ unlabeled examples. In Figure~\ref{fig:2moons}, the examples from $S$ are represented by~\redplus and~\greenminus, and the examples from $T$ are represented by black dots.

We study the adaptation capability of DANN by comparing it to the standard NN.
In our experiments, both algorithms share the same network architecture, with a hidden layer size of $15$ neurons. We even train NN using the same procedure as DANN. That is, we keep updating the domain regressor component using target sample $T$ (with a hyper-parameter $\lambda=6$; the same value used for DANN), but we disable the \emph{adversarial} back-propagation into the hidden layer. To do so, we execute Algorithm~\ref{alg:stoch-up} by omitting the lines numbered~\ref{algoline:omit1} and~\ref{algoline:omit2}. In this way, we obtain a NN learning algorithm -- based on the source risk minimization of Equation~\eqref{eq:loss} -- and simultaneously train the domain regressor of Equation~\eqref{eq:o} to discriminate between source and target domains. Using this toy experiment, we will first illustrate how DANN adapts its decision boundary compared to NN.
Moreover, we will also illustrate how the representation given by the hidden layer is less adapted to the domain task with DANN than it is with NN. 
The results are illustrated in Figure~\ref{fig:2moons}, where the graphs in part (a) relate to the standard NN, and the graphs in part (b) relate to DANN. By looking at the corresponding (a) and (b) graphs in each column, we compare NN and DANN from four different perspectives, described in detail below.

\textbf{Label classification.}\,
The first column of Figure~\ref{fig:2moons} shows the decision boundaries of DANN and NN on the problem of predicting the labels of both source and the target examples. 
As expected, NN  accurately classifies the two classes of the source sample $S$, but is  \emph{not fully adapted} to the target sample $T$.
On the contrary, the decision boundary of DANN perfectly classifies examples from both source and target samples. DANN clearly adapts here to the target distribution.

\textbf{Representation PCA.}\,
To analyze how he domain adaptation regularizer affects the representation $\hh(\cdot)$ provided by the hidden layer, the
second column of Figure~\ref{fig:2moons} presents a principal component analysis (PCA) on the set of all representations of source and target data points, \ie, $\hh(S)\cup\hh(T)$. Thus, given the trained network (NN or DANN), every point from $S$ and~$T$ is mapped into a $15$-dimensional feature space through the hidden layer, and projected back into a two-dimensional plane defined by the first two principal components. 
In the DANN-PCA representation, we observe that target points are homogeneously spread out among the source points. In the NN-PCA representation, clusters of target points containing very few source points are clearly visible. Hence, the task of labeling the target points seems easier to perform on the DANN-PCA representation.

To push the analysis further, four crucial data points identified by A, B, C and D in the graphs of the first column (which correspond to the moon extremities in the original space) are represented again on the graphs of the second column. We observe that points A and B are very close to each other in the NN-PCA representation, while they clearly belong to different classes. The same happens to points C and D. Conversely, these four points are located at opposite corners in the DANN-PCA representation. Note also that the target point A (resp. D) -- which is difficult to classify in the original space -- is located in the~\redplus cluster (resp.~\greenminus cluster) in the DANN-PCA representation.
Therefore, the representation promoted by DANN is more suited for the domain adaptation task.

\textbf{Domain classification.}\,
The third column of Figure~\ref{fig:2moons} shows the decision boundary on the domain classification problem, which is given by  the  domain regressor $o(\cdot)$ of Equation~\eqref{eq:o}. More precisely, $\xb$ is classified as a source example when $o(\xb)\geq 0.5$, and is classified as a domain example otherwise. Remember that, during the learning process of DANN, the $o(\cdot)$ regressor struggles to discriminate between source and target domains, while the hidden representation $\hh(\cdot)$ is \emph{adversarially} updated to prevent it to succeed. As explained above, we trained the domain regressor during the learning process of NN, but without allowing it to influence the learned representation $\hh(\cdot)$.

On one hand, the DANN domain regressor utterly fails to discriminate the source and target distributions. On the other hand, the NN domain regressor shows a better (although imperfect) discriminant. %
This again corroborates that the DANN representation doesn't allow discriminating between domains.

\textbf{Hidden neurons. }
In the plot of the last column of Figure~\ref{fig:2moons}, the lines show the decision surfaces of the hidden layer neurons 
(defined by Equation~\eqref{eq:hh}). In other words, each of the fifteen plot line corresponds to the points $\xb\in\Rbb^2$ for which the $i$th component of $\hh(\xb)$  equals~$\frac{1}{2}$, for $i\in\{1,\ldots,15\}$. 

We observe that the neurons of NN are grouped in three clusters, each one allowing to generate a straight line part of the curved decision boundary for the label classification problem. However, most of these neurons are also able to (roughly) capture the rotation angle of the domain classification problem. Hence, we observe that the adaptation regularizer of DANN prevents these kinds of neurons to be produced. It is indeed striking to see that the two predominant patterns in the NN neurons (\ie, the two parallel lines crossing the plane from the lower left corner to the upper right corner) are absent among DANN neurons.

\subsection{Sentiment Analysis Dataset} 
\label{section:expe_original_data}

\begin{table*}[t]
\centering
\caption{Error rates on the Amazon reviews dataset (left), and Pairwise Poisson binomial test (right).
\label{table:risks} }
\subtable[Error rates on the Amazon reviews dataset]{\small\rowcolors{5}{}{black!10}
\begin{tabular}{lcccccc}
\toprule
\multicolumn{1}{c}{  } & \multicolumn{3}{c}{\textbf{Original data}}  & \multicolumn{3}{c}{\textbf{mSDA representation}} \\
 \cmidrule(l r){2-4} \cmidrule(l r){5-7}
{\it source name $\to$ target name}  & DANN & NN  & SVM & DANN &  NN &  SVM  \\
\cmidrule(l r){1-1} \cmidrule(l r){2-4} \cmidrule(l r){5-7}
 
books $\to$ dvd        & 0.201 & \textbf{0.199} & 0.206 & 0.176   &\textbf{0.171}  & 0.175   \\ 

books $\to$ electronics       &\textbf{0.246}  &0.251  &0.256  &\textbf{0.197}    & 0.228 &  0.244 \\

books $\to$ kitchen           &  0.230  & 0.235 & \textbf{0.229}   &  0.169   &\textbf{0.166} & 0.172 \\ 

dvd $\to$ books                 &  \textbf{0.247} &0.261  & 0.269  & 0.176  & \textbf{0.173} & 0.176 \\ 

dvd $\to$ electronics            &\textbf{0.247}  & 0.256  & 0.249 &\textbf{0.181} & 0.234  & 0.220 \\ 

dvd $\to$ kitchen            & \textbf{0.227}   & \textbf{0.227}  & 0.233  & \textbf{0.151}  & 0.153  & 0.178 \\ 

electronics $\to$ books             & \textbf{0.280} & 0.281  & 0.290  & 0.237 &0.241 & \textbf{0.229} \\

electronics $\to$ dvd             & \textbf{0.273} & 0.277  & 0.278  &\textbf{0.216}& 0.228 &0.261 \\

electronics $\to$ kitchen             & \textbf{0.148}  &0.149  &0.163 &\textbf{0.118}  & 0.126  & 0.137 \\

kitchen $\to$ books             & \textbf{0.283}  & 0.288  &0.325 & \textbf{0.222} & 0.226  &0.234\\

kitchen $\to$ dvd             & \textbf{0.261}  & \textbf{0.261} & 0.274  &  \textbf{0.208}  & 0.214  & 0.209  \\ 

kitchen $\to$ electronics            & 0.161 & 0.161  & \textbf{0.158}  & 0.141 & \textbf{0.136} & 0.138\\  

\bottomrule
\end{tabular}
\label{table:risks_a}} \qquad
\subtable[Pairwise Poisson binomial test]{\small\begin{tabular}{lccc}
\toprule
 \multicolumn{4}{c}{\textbf{Original data}}  \\
 \midrule
 & DANN & NN  & SVM   \\[1mm]
DANN        & 0.50 &  \textbf{0.90} & \textbf{0.97}   \\[1mm]

NN           & 0.10 & 0.50 & 0.87   \\[1mm]

SVM            & 0.03 &  0.13 & 0.50  \\
\bottomrule\\[2mm]
\toprule
 \multicolumn{4}{c}{\textbf{mSDA representations}}  \\
 \midrule
 & DANN & NN  & SVM   \\[1mm]
 
DANN      &    0.50 &  \textbf{0.82}  & \textbf{0.88}  \\[1mm]

NN       &     0.18 & 0.50 & 0.90 \\ [1mm]

SVM            & 0.12 & 0.10 &  0.50 \\
\bottomrule
\end{tabular}
\label{table:risks_b}}
\end{table*}
In this section, we compare the performance of our proposed DANN algorithm to a standard neural network with one hidden layer (NN) described by Equation~\eqref{eq:loss}, and a Support Vector Machine (SVM) with a linear kernel. To select the hyper-parameters of each of these algorithms, we use grid search and a very small validation set which consists in $100$ labeled examples from the target domain. Finally, we select the classifiers having the lowest target validation risk. 

We compare the algorithms on the {\it Amazon reviews} dataset, as pre-processed by~\citet{Chen12}. 
This dataset includes four domains, each one composed of reviews of a specific kind of product (books, dvd disks, electronics, and kitchen appliances). Reviews are encoded in $5\,000$ dimensional feature vectors of unigrams and bigrams, and labels are binary: ``$0$'' if the product is ranked up to $3$ stars, and ``$1$'' if the product is ranked $4$ or $5$ stars.

We perform twelve domain adaptation tasks. For example, ``books $\to$ dvd'' corresponds to the task for which books is the source domain and dvd disks the target one.
All learning algorithms are given  $2\,000$ labeled source examples and $2\,000$ unlabeled target examples. Then, we evaluate them on separate target test sets (between $3\,000$ and $6\,000$ examples).
Note that NN and SVM don't use the unlabeled target sample for learning. Here are more details about the procedure used for each learning algorithms.
 
\textbf{DANN. } The adaptation parameter $\lambda$ is chosen among 9 values between $10^{-2}$ and $1$ on a logarithmic scale. The hidden layer size $l$ is either $1, 5, 12, 25, 50, 75, 100, 150,$ or~$200$. Finally, the learning rate $\alpha$ is fixed at $10^{-3}$.

\textbf{NN. }  We use exactly the same hyper-parameters and training procedure as DANN above, except that we don't need an adaptation parameter. Note that one can train NN by using the DANN implementation (Algorithm~\ref{alg:stoch-up}) with $\lambda=0$.

\textbf{SVM. } The hyper-parameter $C$ of the SVM  is chosen among 10 values between $10^{-5}$ and $1$ on a logarithmic scale. This range of values is the same used by~\citet{Chen12} in their experiments.

The ``Original data'' part of Table~\ref{table:risks_a} shows
the target test risk of all algorithms, and Table~\ref{table:risks_b}
reports the probability that one algorithm is significantly better than another according to the Poisson binomial test~\cite{lacoste-2012}.
We note that DANN has a significantly better performance than NN and SVM, with respective probabilities \textbf{0.90} and~\textbf{0.97}.  As the only difference between DANN and NN is the domain adaptation 
regularizer, we conclude that our approach successfully helps to find a representation suitable for the target domain.

\begin{figure*}[t]
\centering
\subfigure[DANN on \emph{Original data}.]{
\includegraphics[width=.315\textwidth]{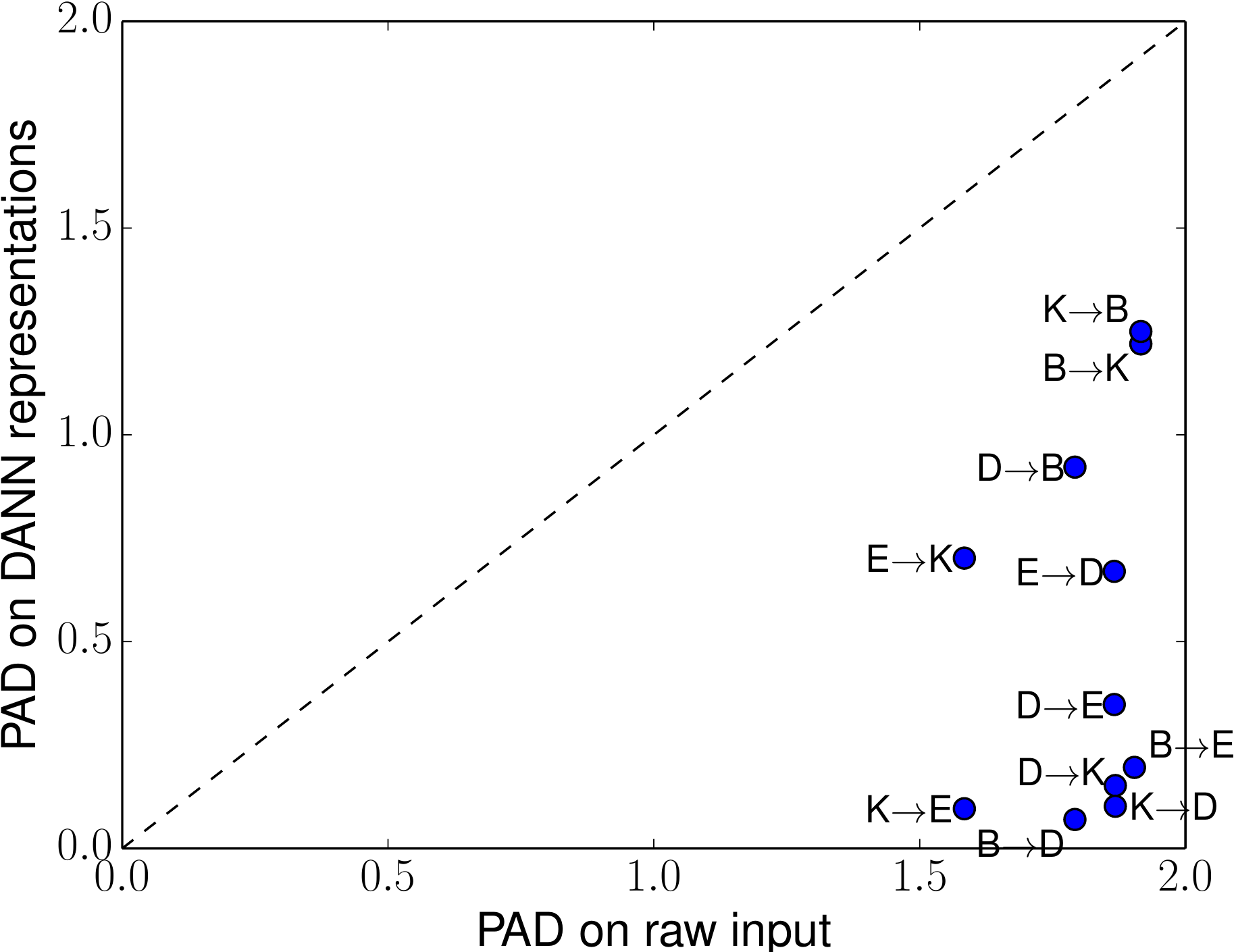}
\label{fig:PAD_a}
}\ \ 
\subfigure[DANN \& NN with  100 hidden neurons.]{
\includegraphics[width=.31\textwidth]
{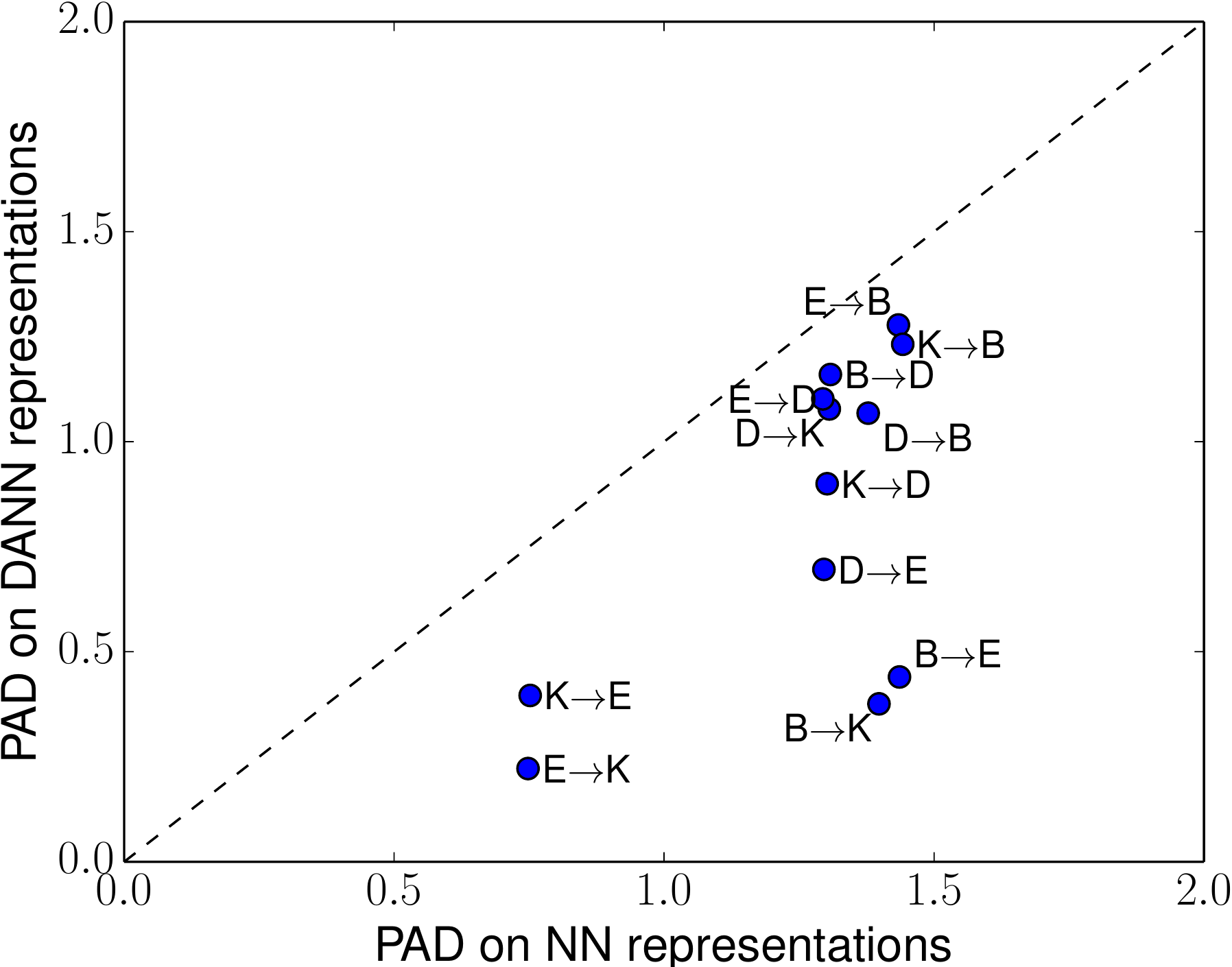}
\label{fig:PAD_b}
}\ \ 
\subfigure[DANN on \emph{mSDA representations.}]{
\includegraphics[width=.315\textwidth]
{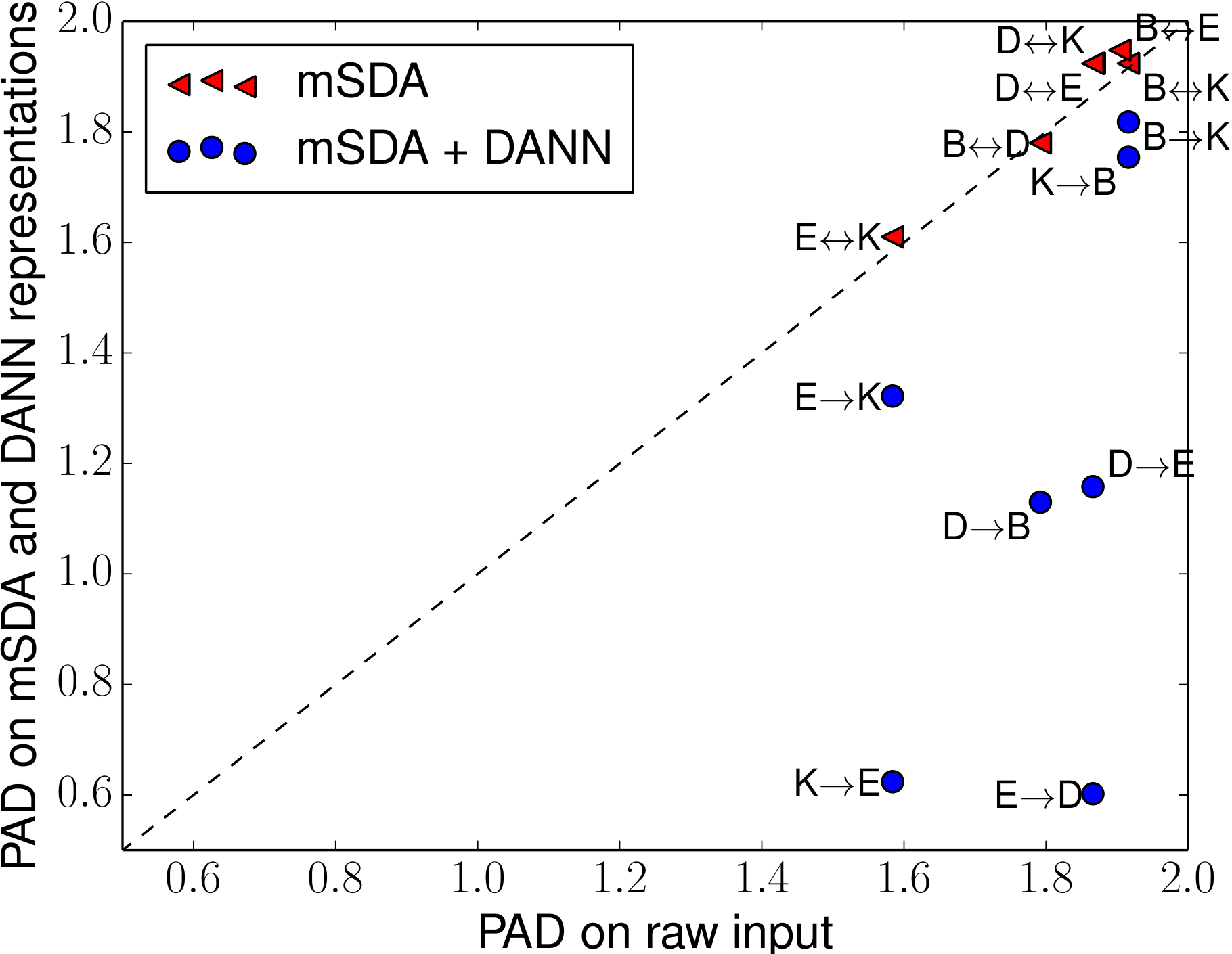}
\label{fig:PAD_c}
}
\caption{Proxy A-distances (PAD).  Note that the PAD values of mSDA representations are symmetric when swapping source and target samples. Also, some DANN results, when used on top of mSDA, have PAD values lower than $0.5$ and doesn't appear on Figure~(c).}
\end{figure*}

\subsection{Combining DANN with Autoencoders}
\label{section:autoencoders}
 
We now wonder whether our DANN algorithm can improve on the representation learned by the state-of-the-art \emph{Marginalized Stacked Denoising Autoencoders} (mSDA) proposed by~\citet{Chen12}.
In brief, mSDA is an unsupervised algorithm that learns a new  robust feature representation of the training samples. 
It takes the unlabeled parts of both source and target samples 
to learn a feature map from the input space $\Xcal$ to a new representation space. 
As a \emph{denoising autoencoder}, it finds a feature representation from which one can (approximately) reconstruct
the original features of an example from its noisy counterpart.
\citet{Chen12} showed that using mSDA with a linear SVM classifier gives state-of-the-art performance on the \emph{Amazon reviews} datasets. As an alternative to the SVM, we propose to apply our DANN algorithm on the same representations generated by mSDA (using representations of both source and target samples). Note that, even if mSDA and DANN are two representation learning approaches, they optimize different objectives, which can be complementary. 

We perform this experiment on the same \emph{amazon reviews} dataset described in the previous subsection. For each pair source-target, we generate the mSDA representations using a corruption probability of~$50\%$ and a number of layers of~$5$.  We then execute the three learning algorithms (DANN, NN, and SVM) on these representations. 
More precisely, following the experimental procedure of \citet{Chen12}, we use the concatenation of the output of the~$5$ layers and the original input as the new representation. Thus, each example is now encoded in a vector of $30\,000$ dimensions.
Note that we use the same grid search as in Subsection~\ref{section:expe_original_data}, but with a learning rate $\alpha$ of $10^{-4}$ for both DANN and NN.
The results of ``mSDA representation'' columns in Table~\ref{table:risks_a}  confirm that combining mSDA and DANN is a sound approach. Indeed, the Poisson binomial test shows that DANN has a better performance than NN and SVM with probabilities \textbf{0.82} and \textbf{0.88} respectively, as reported in Table~\ref{table:risks_b}.

\subsection{Proxy A-distance} 

The theoretical foundation of DANN is the domain adaptation theory of~\citet{BenDavid-NIPS06,BenDavid-MLJ2010}. We claimed that DANN finds a representation in which the source and the target example are hardly distinguishable. Our toy experiment of Section~\ref{sec:toy_problem} already points out some evidences, but we want to confirm it on real data. To do so, we compare the Proxy A-distance (PAD) on various representations of the \emph{Amazon Reviews} dataset. These representations are obtained by running either NN, DANN, mSDA, or mSDA and DANN combined. Recall that PAD, as described in Section~\ref{section:PAD}, is a metric estimating the similarity of the source and the target representations. 
More precisely, to obtain a PAD value, we use the following procedure: (1) we construct the dataset $U$ of Equation~\eqref{eq:U} using both source and target representations of the training samples; (2) we randomly split $U$ in two subsets of equal size; (3) we train linear SVMs on the first subset of~$U$ using a large range of $C$ values; (4) we compute the error of all obtained classifiers on the second subset of~$U$; and (5) we use the lowest error to compute the PAD value of Equation~\eqref{eq:PAD}.

Firstly, Figure~\ref{fig:PAD_a} compares the PAD of DANN representations obtained in the experiments of Section~\ref{section:expe_original_data} (using the hyper-parameters values leading to the results of Table~\ref{table:risks}) to the PAD computed on raw data. As expected, the PAD values are driven down by the DANN representations. 

Secondly, Figure~\ref{fig:PAD_b} compares the PAD of DANN representations to the PAD of standard NN representations.
As the PAD is influenced by the hidden layer size (the discriminating power tends to increase with the dimension of the representation), we fix here the size to $100$ neurons for both algorithms. We also fix the adaptation parameter of DANN to $\lambda \simeq 0.31$ as it was the value that has been selected most of the time during our preceding experiments on the \emph{Amazon Reviews} dataset. Again, DANN is clearly leading to the lowest PAD values.  

Lastly, Figure~\ref{fig:PAD_c} presents two sets of results related to Section~\ref{section:autoencoders} experiments. On one hand, we reproduce the results of \citet{Chen12}, which noticed that the mSDA representations gave greater PAD values than those obtained with the original (raw) data. Although the mSDA approach clearly helps to adapt to the target task, it seems to contradict the theory of \citeauthor{BenDavid-NIPS06}. On the other hand, we observe that, when running DANN on top of mSDA (using the hyper-parameters values leading to the results of Table~\ref{table:risks}), the obtained representations have much lower PAD values. These observations might explain the improvements provided by DANN when combined with the mSDA procedure.

\section{Conclusion and Future Work}
In this paper, we have proposed a neural network algorithm, named DANN, that is strongly inspired by the domain adaptation theory of \citet{BenDavid-NIPS06,BenDavid-MLJ2010}. 
The main idea behind DANN is to encourage the network's hidden layer to learn a representation which is predictive of the source example labels, but uninformative about the domain of the input (source or target). Extensive experiments on the \emph{inter-twinning moons} toy problem and \emph{Amazon reviews} sentiment analysis dataset have shown the effectiveness of this strategy. Notably, we achieved state-of-the-art performances when combining DANN with the mSDA autoencoders of \citet{Chen12}, which turned out to be two complementary representation learning approaches.

We believe that our domain adaptation regularizer that we develop for the DANN algorithm can be incorporated into many other learning algorithms.  Natural extensions of our work would be deeper network architectures, multi-source adaptation problems and other learning tasks beyond the basic binary classification setting. We also intend to meld the DANN approach with denoising autoencoders, to potentially improve on the two steps procedure of Section~\ref{section:autoencoders}.

\bibliography{biblio}
\bibliographystyle{icml2015}

\end{document}